\documentclass[letterpaper]{article} 
\usepackage{aaai19m}  
\usepackage{times}  
\usepackage{helvet} 
\usepackage{courier}  
\usepackage[hyphens]{url}  
\usepackage{graphicx} 
\usepackage{graphicx}  
\usepackage{amsmath}
\usepackage{amssymb}
\usepackage{color}

\usepackage{subfigure}
\usepackage{booktabs}
\usepackage{multirow}
\usepackage{diagbox}
\def\ie{{\em i.e.}}

\def\etal{{\em et al.}}

\title{PedHunter: Occlusion Robust Pedestrian Detector in Crowded Scenes}
\author{Cheng Chi$^{1,3*}$, Shifeng Zhang$^{2,3}$\thanks{These authors contributed equally to this work.}, Junliang Xing$^{2,3}$, Zhen Lei$^{2,3}$, Stan Z. Li$^{2,3}$, Xudong Zou$^{1,3}$\\
$^{1}$ Institute of Electronics, Chinese Academy of Sciences, Beijing, China\\
$^{2}$CBSR \& NLPR, Institute of Automation, Chinese Academy of Sciences, Beijing, China\\
$^{3}$University of Chinese Academy of Sciences, Beijing, China
}
 
\begin{document}

\maketitle

\begin{abstract}
Pedestrian detection in crowded scenes is a challenging problem, because occlusion happens frequently among different pedestrians. In this paper, we propose an effective and efficient detection network to hunt pedestrians in crowd scenes. The proposed method, namely PedHunter, introduces strong occlusion handling ability to existing region-based detection networks without bringing extra computations in the inference stage. Specifically, we design a mask-guided module to leverage the head information to enhance the feature representation learning of the backbone network. Moreover, we develop a strict classification criterion by improving the quality of positive samples during training to eliminate common false positives of pedestrian detection in crowded scenes. Besides, we present an occlusion-simulated data augmentation to enrich the pattern and quantity of occlusion samples to improve the occlusion robustness. As a consequent, we achieve state-of-the-art results on three pedestrian detection datasets including CityPersons, Caltech-USA and CrowdHuman. To facilitate further studies on the occluded pedestrian detection in surveillance scenes, we release a new pedestrian dataset, called SUR-PED, with a total of over $162k$ high-quality manually labeled instances in $10k$ images. The proposed dataset, source codes and trained models will be released.
\end{abstract}

\section{Introduction}
Pedestrian detection is an important research topic in computer vision with various applications, such as autonomous driving, video surveillance and robotics. The goal of pedestrian detection is to predict a bounding box for each pedestrian instance in an image. Although great progress has been made in the realm of pedestrian detection \cite{DBLP:conf/cvpr/TianLWT15,DBLP:conf/cvpr/ZhangBS15,DBLP:conf/cvpr/CosteaN16,DBLP:conf/wacv/DuELD17,Lin_2018_ECCV,Noh_2018_CVPR}, occlusion still remains as one of the most challenging issues. Some efforts have been made particularly to handle the occlusion problem, but a notable amount of extra computations is introduced and there is still much room to improve the detection performance.

In this work, we propose a high-performance pedestrian detector namely PedHunter to improve occlusion robustness in crowded scenes, which mitigates the impact of occlusion without sacrificing inference speed. Firstly, inspired by human visual system that often resorts to the head information to locate each pedestrian in crowded scenes, we propose a mask-guided module to predict head masks that helps to enhance the representation learning of pedestrian features during training. The mask-guided module does not participate in inference, so it improves the accuracy without computational overhead. Secondly, we observe that error detections between occluded pedestrians are the common false positive in crowd scenes, which results from the low-quality positive samples during training. To improve the quality of positive samples, we develop a strict classification criterion by increasing the IoU threshold of positive samples and adding jittered ground truths, which reduces aforementioned false positives. Thirdly, we introduce an occlusion-simulated data augmentation technique that generates random occlusion on ground truths during training. It significantly enriches the pattern and quantity of occlusion, making the proposed model more robust to occlusion.

In addition, current pedestrian detection benchmarks usually focus on autonomous driving scenes. It is necessary to build a new benchmark in the field of surveillance, which is another important application domain for pedestrian detection. To this end, we introduce a new SUR-PED dataset for occluded pedestrian detection in surveillance scenes. It has over $162k$ pedestrian instances manually labeled in $10k$ surveillance camera images. Based on this proposed dataset and exiting datasets including CrowdHuman \cite{DBLP:journals/corr/abs-1805-00123}, extented CityPersons \cite{DBLP:conf/cvpr/ZhangBS17} and extented Caltech-USA \cite{DBLP:conf/cvpr/DollarWSP09}\footnote{In our previous work, we additionally label the corresponding head bounding box for each annotated pedestrian instance as the extended version of CityPersons and Caltech-USA.}, several experiments are conducted to demonstrate the superiority of the proposed method, especially for the crowded scenes. Notably, the proposed PedHunter detector achieves state-of-the-art results without adding any extra overhead, \ie, $8.32\%$ MR$^{-2}$ on CityPersons, $2.31\%$ MR$^{-2}$ on Caltech-USA and $39.5\%$ MR$^{-2}$ on CrowdHuman.

To summarize, the main contributions of this work are in five-fold as follows: 1) Proposing a mask-guided module to enhance the discrimination ability of the occluded features; 2) Designing a strict classification criterion to provide higher quality positives for R-CNN to reduce false positives; 3) Developing an occlusion-simulated data augmentation to enrich occlusion diversity for stronger robustness; 4) Providing a new surveillance pedestrian dataset to facilitate further studies on occluded pedestrian detection; 5) Achieving state-of-the-art results on common pedestrian detection datasets without adding additional overhead.

\begin{figure*}[t]
\centering
\includegraphics[width=0.95\linewidth]{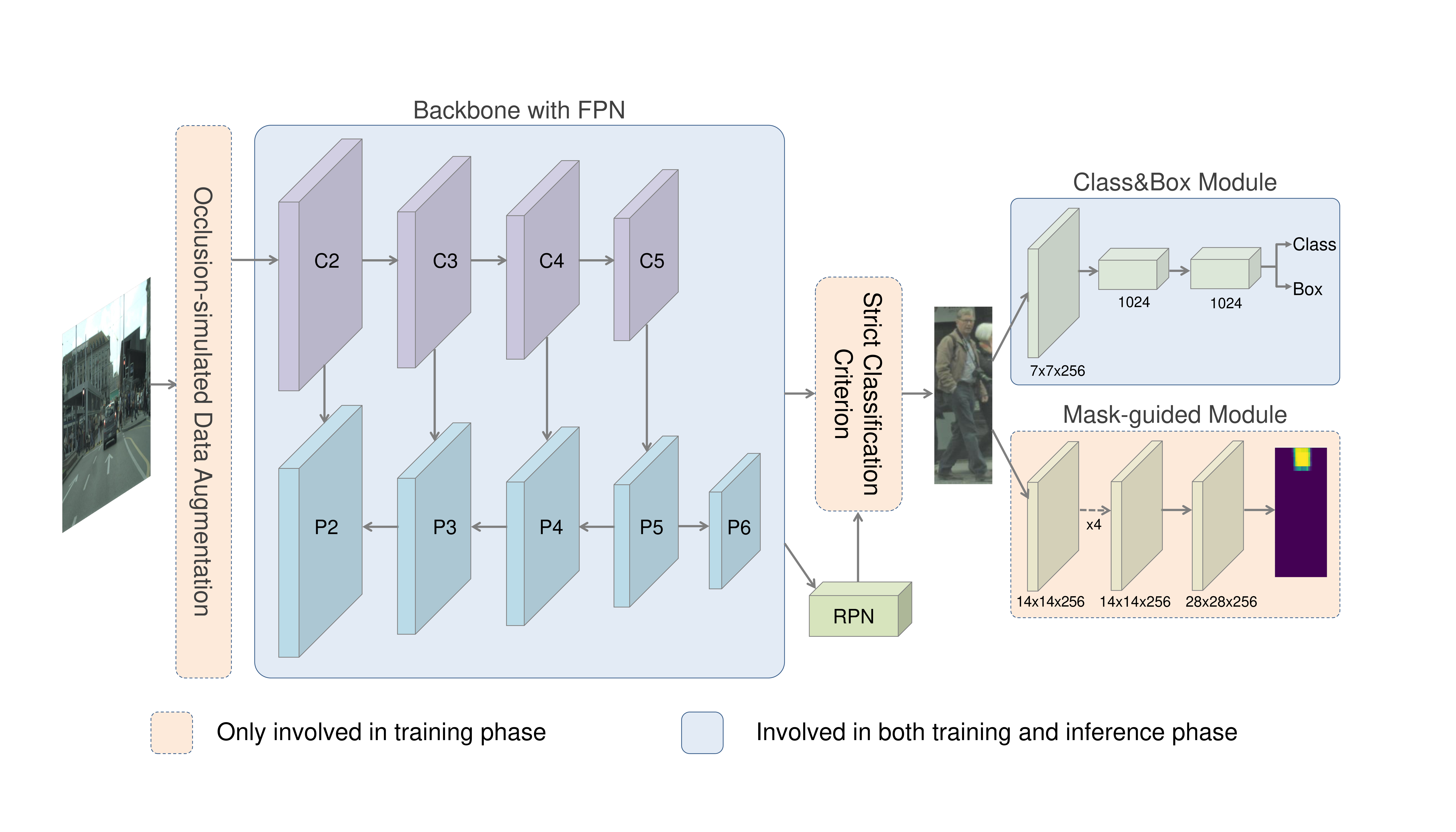}
\caption{Structure of PedHunter. It contains Backbone with FPN, RPN, Class\&Box and Mask-guided Module. The proposed components are shown within orange dotted rectangle. Through Occlusion-simulated Data Augmentation, input images with random occlusion are sent into backbone and then RPN generates candidate bounding boxes. After that, we use strict classification criterion to provide higher-quality positives to subsequent modules. The Class\&Box Module performs the second-stage classification and regression, and the Mask-guided Module predicts the associating head mask.}
\label{fig:framework}
\end{figure*}

\section{Related Work}

Pedestrian detection is dominated by CNN-based methods \cite{DBLP:conf/cvpr/HosangOBS15,DBLP:conf/iccv/YangYLL15} in recent years. Sermanet \etal~\cite{DBLP:conf/cvpr/SermanetKCL13} use the convolutional sparse coding to pre-train CNN for pedestrian detection. Cai \etal~\cite{DBLP:conf/iccv/CaiSV15} persent a complexity-aware cascaded detector for an optimal trade-off between accuracy and speed. Angelova \etal~\cite{DBLP:conf/bmvc/AngelovaKVOF15} detect pedestrian by combining the ideas of fast cascade and a deep network. Zhang \etal~\cite{DBLP:conf/eccv/ZhangLLH16} present an effective pipeline for pedestrian detection via using RPN followed by boosted forests. Mao \etal~\cite{DBLP:conf/cvpr/MaoXJC17} introduce a novel network architecture to jointly learn pedestrian detection with the given extra features. Li \etal~\cite{DBLP:journals/tmm/LiLSXFY18} use multiple built-in sub-networks to adaptively detect pedestrians across scales. Brazil \etal~\cite{DBLP:conf/iccv/BrazilYL17} exploit weakly annotated boxes via a segmentation infusion network to achieve considerable performance gains.

Although significant progresses have been made from CNN-based pedestrian methods, it remains a very challenging problem to detect occluded pedestrian in crowd scenes. Several methods \cite{DBLP:conf/iccv/TianLWT15} describe the pedestrian using part-based model to handle occlusion, which learn a series of part detectors and design some mechanisms to fuse the part detection results to localize partially occluded pedestrians. Besides the part-based model, Zhou \etal~\cite{DBLP:conf/iccv/ZhouY17} present to jointly learn part detectors to exploit part correlations as well as reduce the computational cost. Wang \etal~\cite{DBLP:conf/cvpr/WangXJSSS18} propose a novel bounding box regression loss to detect pedestrians in the crowd scenes. Zhang \etal~\cite{DBLP:conf/cvpr/Zhang0S18} propose to utilize channel-wise attention in convnets allowing the network to learn more representative features for different occlusion patterns in one coherent model. Zhang \etal~\cite{DBLP:conf/eccv/ZhangWBLL18} design an aggregation loss to enforce proposals to be close and locate compactly to the corresponding objects. Zhou \etal~\cite{DBLP:conf/eccv/ZhouY18} design a method to detect full body and visible part estimation simultaneously to further estimate occlusion. Although numerous pedestrian detection methods are presented, how to effectively detect each individual pedestrian in crowded scenarios is still one of the most critical issues for pedestrian detectors.

Behind those different methods, there are several datasets \cite{DBLP:conf/cvpr/DalalT05,DBLP:conf/iccv/EssLG07,geronimo2007adaptive,overett2008new,DBLP:conf/ivs/SilbersteinLKG14,DBLP:conf/cvpr/WojekWS09,DBLP:conf/iccv/WuN07} that provide strong support for pedestrian detection in the last decade. The Tsinghua-Daimler Cyclist (TDC)~\cite{DBLP:conf/ivs/LiFYXBPLG16} dataset focuses on cyclists recorded from a vehicle-mounted stereo vision camera, containing a large number of cyclists varying widely in appearance, pose, scale, occlusion and viewpoint. The KITTI~\cite{DBLP:conf/cvpr/GeigerLU12} dataset focuses on autonomous driving and is collected via a standard station wagon with two high-resolution color and grayscale video cameras, around the mid-size city of Karlsruhe, in rural areas and on highways, up to $15$ cars and $30$ pedestrians are visible per image. The EuroCity Persons dataset~\cite{DBLP:journals/corr/abs-1805-07193} provides a large number of highly diverse, accurate and detailed annotations of pedestrians, cyclists and other riders in $31$ cities of $12$ different European countries.

\begin{figure*}[h]
\centering
\subfigure[Input image]{
\label{fig:input} 
\includegraphics[width=0.31\linewidth]{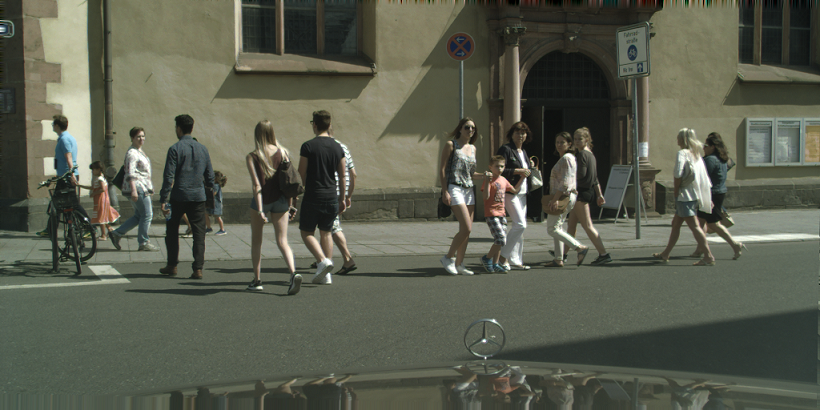}}
\subfigure[Baseline]{
\label{fig:fpn} 
\includegraphics[width=0.31\linewidth]{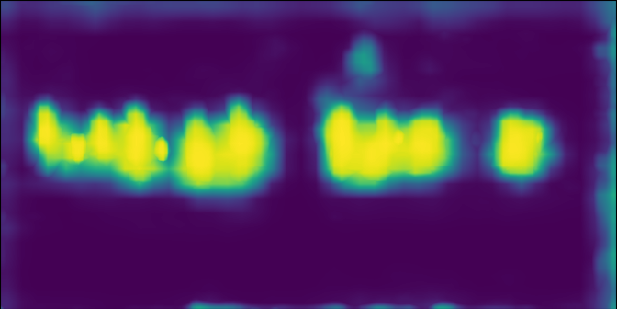}}
\subfigure[PedHunter]{
\label{fig:ours} 
\includegraphics[width=0.31\linewidth]{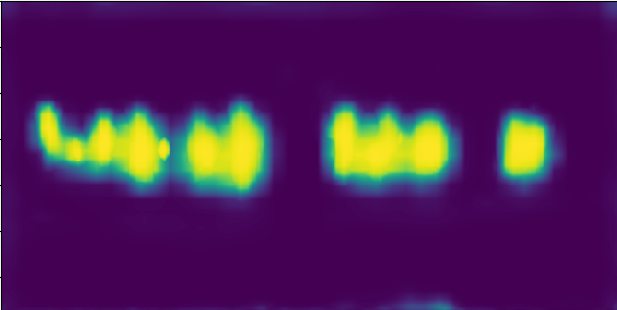}}
\vspace{-3.0mm}
\caption{Proposal score feature map visualization. Compared to the baseline, PedHunter shows stronger response to pedestrians, \ie, the color of pedestrian region in (c) is brighter than (b). And the pedestrian boundary on feature map of PedHunter is more clear than baseline, indicating that adjacent pedestrians are more distinguishable in our method.}
\label{fig:feat-vis}
\end{figure*}

\section{PedHunter}
Figure \ref{fig:framework} shows the architecture of PedHunter, which can adopt any existing networks as the backbone. In this work, we use ResNet-50 \cite{DBLP:conf/cvpr/HeZRS16} with 5-level feature pyramid structure as a demonstration. On the basis of FPN \cite{DBLP:conf/cvpr/LinDGHHB17} baseline, we propose three zero-cost components to enhance its occlusion handling ability, \ie, the mask-guided module, the strict classification criterion, and the occlusion-simulated data augmentation. Each component is described below.

\subsection{Mask-guided Module}
\label{mask}

It is usually difficult for CNN-based methods to detect occluded pedestrians in crowded scenes, because pedestrians often occlude each other and their features become intertwined after several convolution layers. Thus, more discriminative features of occluded pedestrians need to be learned to hunt these targets in crowded scenes. Compared with other body parts, head is more visible, the occlusion pattern statistics in CityPersons also confirm this. Since head is more visible, people often resort to head to find corresponding pedestrian in crowd scenes. Inspired by this observation, the mask-guided module is designed to utilize the head location as an extra supervision, which assists the network to learn more discriminative features for occluded pedestrians. As shown in Figure \ref{fig:framework}, this newly added module is in parallel with the existing class\&bbox module to predict head masks. In particular, it utilizes four stacked $3\times3$ conv layers, one deconv layer, one $1\times1$ conv layer and a per-pixel sigmoid function to predict a binary head mask for each pedestrian proposal. We adopt the box-wise annotation as the segmentation supervision of the proposed module for simplicity. The average binary cross-entropy loss is applied to train this module. 

Through the supervision of head segmentation information, the learned features of occluded pedestrians tend to be more discriminative. The feature map visualization of baseline and the proposed PedHunter in Figure \ref{fig:feat-vis} also illustrates the enhancement brought by the mask-guided module. Compared to the baseline model, the learned features of PedHunter show stronger response to pedestrians and the distinction of adjacent pedestrians is more obvious. Besides, the proposed module only works in the training phase, so that the detector can keep the same computational cost as the original network during inference.

The mask-guided module can be considered as another form of attention mechanism to assist the feature learning. Compared with the external attention guidance method (marked as GA) in \cite{DBLP:conf/cvpr/Zhang0S18}, our mask-guided module has the following two advantages: 1) Our method does not use extra dataset, while GA uses the MPII Pose dataset to pre-train part detector. 2) During inference, GA also need the external part detector to generate the attention vector, leading to considerable additional computing overhead. However, our method avoids any additional overhead. Besides, we would like to emphasize that the mask-guided module uses head to \textbf{assist the model in learning better features} during training. During inference, even a pedestrian does not have a visible head, the learned better features for this pedestrian can also help the R-CNN branch for better detection performance.

\begin{figure}[t]
\centering
\subfigure[]{
\label{fig:fp} 
\includegraphics[width=0.3\linewidth]{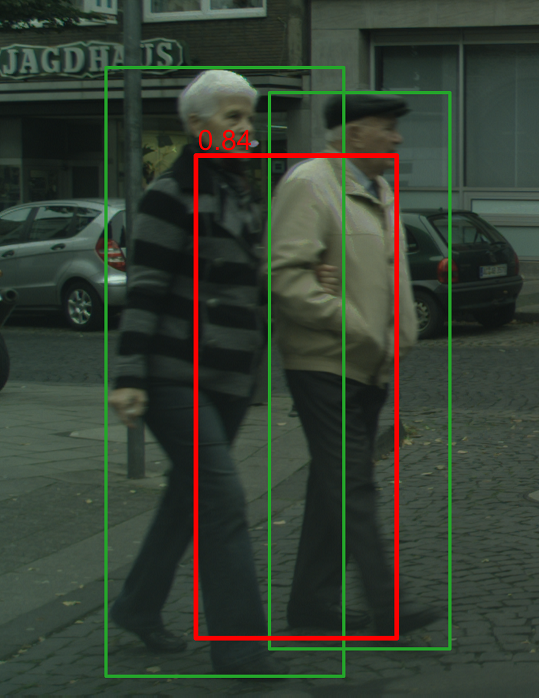}}
\subfigure[]{
\label{fig:pp} 
\includegraphics[width=0.3\linewidth]{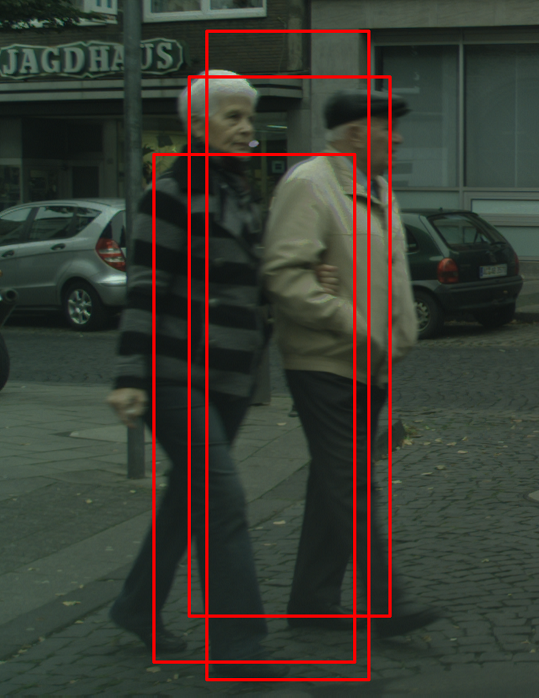}}
\subfigure[]{
\label{fig:gp} 
\includegraphics[width=0.3\linewidth]{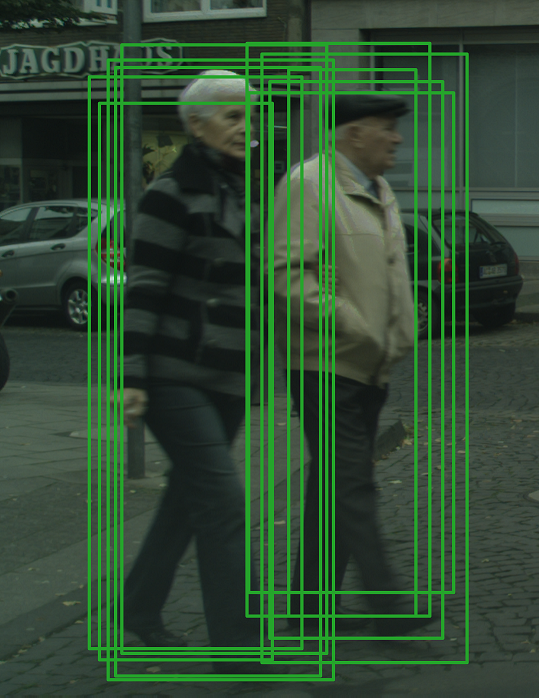}}
\vspace{-3.0mm}
\caption{(a) A false positive with high score between overlapping pedestrians that is not suppressed by true positives. (b) Several poor-quality positive samples in the R-CNN stage. (c) Improved positive samples under the introduced strict classification criterion.}
\label{fig:proposal} 
\end{figure}

\subsection{Strict Classification Criterion}
\label{pis}
Another key problem in occluded pedestrian detection is the false positive between overlapping pedestrians shown in Figure \ref{fig:fp}. These false positives are hard to be suppressed by Non-Maximum Suppression (NMS) because of its sensitivity to the IoU threshold. The main cause for these false positives is the poor quality of positive samples during training R-CNN. As shown in Figure \ref{fig:pp}, RPN often generates some proposals between occluded pedestrians and most of these proposals will be matched as positive samples in R-CNN. During inference, these proposals will be predicted with a high score and thus damage the performance.

To address this issue, we introduce a new strict classification criterion containing the following two steps:

\begin{itemize}
\setlength{\itemsep}{2pt}
\setlength{\parsep}{0pt}
\setlength{\parskip}{0pt}
\item Increasing IoU threshold for positive samples. To improve the quality of positive samples for R-CNN, we increase the IoU threshold of matching positive samples from $0.5$ to $0.7$ in the second stage. Only the proposals with higher overlaps with ground truths are treated as positive samples, while those proposals lying in the union of overlapping pedestrians with lower IoU are treated as ignored or negative samples.
\item Jittering ground truths. The first step increasing IoU threshold will greatly reduce the number of positive samples, causing a serious class imbalance problem. To solve this issue, we introduce a ground truth jittering operation, which randomly jitters each ground truth $10$ times with a small amplitude [$\delta_{x_1}$,$\delta_{y_1}$,$\delta_{x_2}$,$\delta_{y_2}$], where $\delta_{x_1}$,$\delta_{x_2}$ $\negmedspace\sim\negmedspace$ $Uniform$($-0.2w$, $0.2w$) and $\delta_{y_1}$,$\delta_{y_2}$ $\negmedspace\sim\negmedspace$ $Uniform$($-0.2h$, $0.2h$), $w$ and $h$ are width and height of ground truths. Then these jittered boxes are added into proposals to train R-CNN.
\end{itemize}

As shown in Figure \ref{fig:gp}, the proposed criterion is able to remove those poor-quality positive samples shown in Figure \ref{fig:pp} and significantly improves the quality of the positive samples. During the inference phase, those poor-quality proposals are more inclined to get lower confidence scores and have a minor impact on the final performance.

\subsection{Occlusion-simulated Data Augmentation}
Another reason to the relatively poor performance of occluded pedestrian detection is that there is only a small percentage of occlusion cases during training. Taking the CityPersons dataset as an example, more than $50\%$ of the instances only possess less than $20\%$ occlusion ratio. There is fewer occlusion cases in the Caltech-USA dataset.

\begin{figure}[t]
\centering
\includegraphics[width=0.9\linewidth]{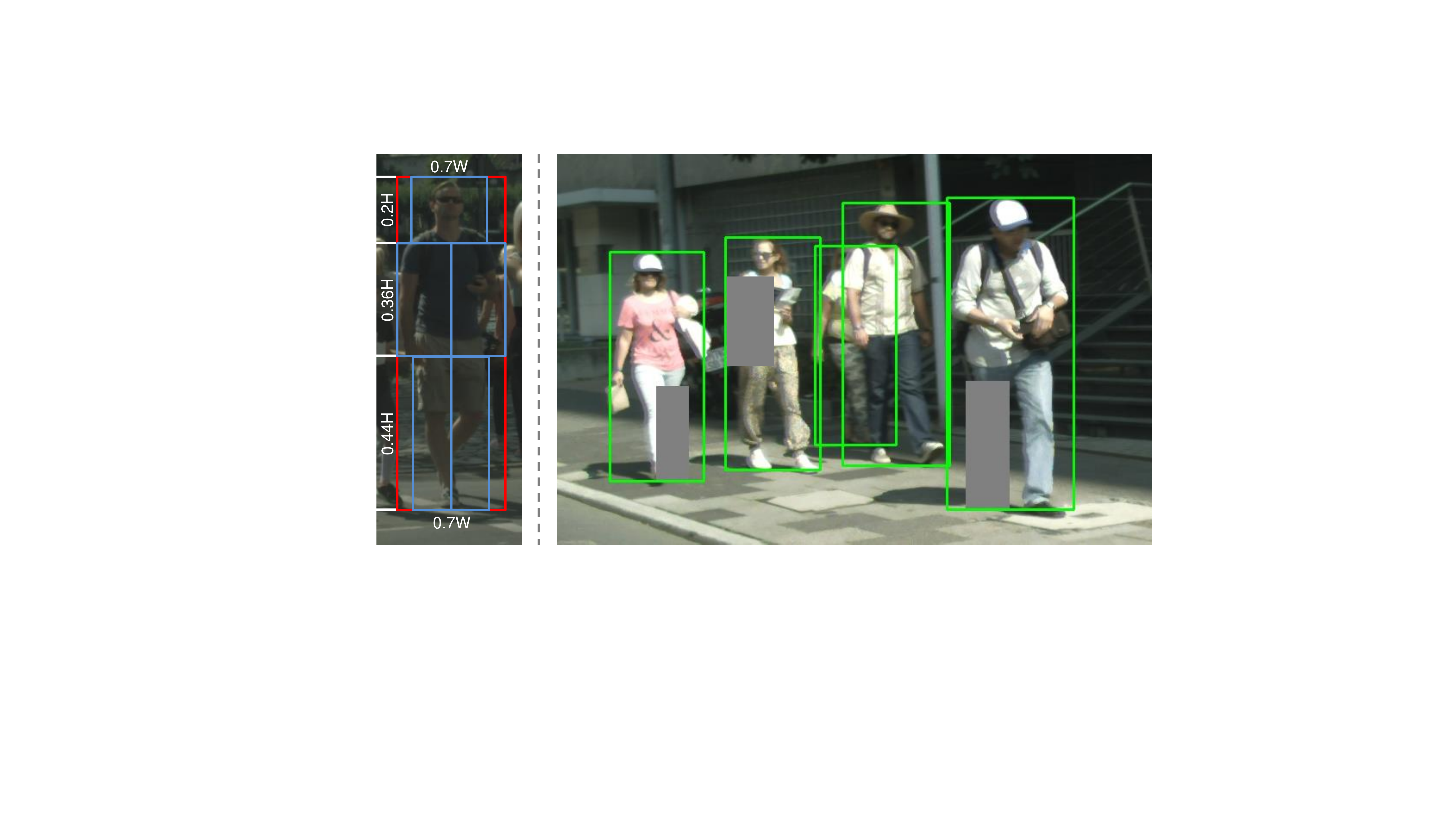}
\vspace{-3.0mm}
\caption{We divide each ground truth into $5$ parts (\ie, head, left upper body, right upper body, left leg and right leg). During training, we randomly select one part except the head to add occlusion.}
\label{fig:occ_aug}
\end{figure}

To diversify the occlusion cases in the training phase, we propose an occlusion-simulated data augmentation to construct a more occlusion-robust model. As shown in Figure \ref{fig:occ_aug}, we first divide each ground truth into five parts (\ie, head, left upper body, right upper body, left leg, and right leg) with the empirical ratio in \cite{DBLP:journals/pami/FelzenszwalbGMR10}, then randomly select one part except the head to occlude with the mean-value of ImageNet \cite{DBLP:journals/ijcv/RussakovskyDSKS15}. This data augmentation is used for each ground truth with a probability of $0.5$ to ensure randomness.
 
With the proposed data augmentation, the quantity and pattern of occlusion cases can be significantly enriched during training. Since $50\%$ pedestrians with $1/5$ body are blanked (\ie, only $10\%$ pedestrian regions are blanked and $90\%$ pedestrian regions are natural), thus the detector will not pay excessive attention on added occlusion regions. Besides, similar to that randomly adding some noises on the image can improve the network robustness, this augmentation randomly adds some occlusion parts to improve the occlusion robustness. The improvements brought by this augmentation (stated in Section Model Analyse) can prove the above statement. Additionally, the method also does not bring extra computational cost during the inference phase.

\begin{figure*}[t]
\centering
\includegraphics[width=0.94\textwidth]{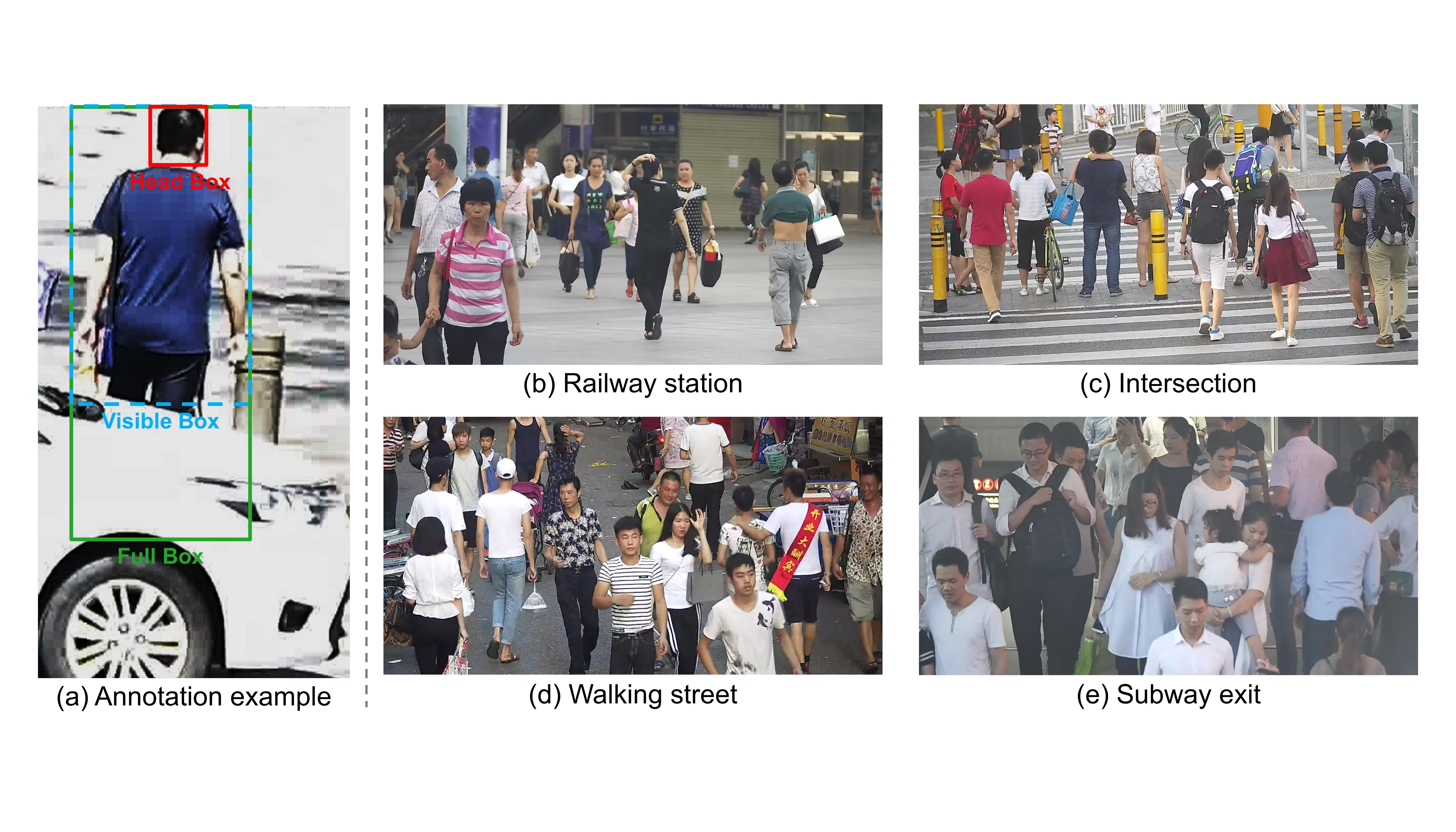}
\caption{(a) An illustrative example of three kinds of annotations: head bounding-box, visible bounding-box and full bounding-box. (b)-(e) Four common scenes in SUR-PED including railway station, intersection, walking street and subway exit.}
\label{fig:sur-ped}
\end{figure*}

\subsection{Training and Inference}

{\flushleft \textbf{Anchor Design.} }
At each location of the detection layer, we only associate one specific scale of anchors (\ie, $8S$, where $S$ represents the downsampling factor of the detection layer). They cover the scale range $32-512$ pixels across different levels with respect to the network's input image. We use different anchor aspect ratios for different datasets, which is described in the next section.

{\flushleft \textbf{Sample Matching.} }
Samples are assigned to ground-truth pedestrian boxes using an IoU threshold of $\theta_{p}$, and to background if their IoU is in $[0, \theta_{n})$. If an anchor is unassigned, which may happen with overlap in $[\theta_{n}, \theta_{p})$, it is ignored during the training phase. Based on the proposed strict classification criterion, we set $\theta_{n}=0.3$ and $\theta_{p}=0.7$ for the RPN stage same as original, and $\theta_{n}=0.5$ and $\theta_{p}=0.7$ for the R-CNN stage. 

{\flushleft \textbf{Loss Function.} }
The whole network is optimized by ${\cal L}={\cal L}_\text{cls} + {\lambda_1} {\cal L}_\text{box} + {\lambda_2} {\cal L}_\text{mask}$, where the classification loss ${\cal L}_\text{cls}$ and the regression loss ${\cal L}_\text{box}$ are identical as those defined in \cite{DBLP:conf/iccv/Girshick15}. The mask loss ${\cal L}_\text{mask}$ is the average binary cross-entropy loss. The loss weight coefficients ${\lambda_1}$ and ${\lambda_2}$ are used to balance different loss terms and we empirically set them as $1$ in all experiments.

{\flushleft \textbf{Optimization.} }
The backbone network is initialized by the ImageNet \cite{DBLP:journals/ijcv/RussakovskyDSKS15} pretrained ResNet-50 model. 
The parameters of newly added layers in the RPN are initialized by the normal distribution method, and the parameters in the class\&box module and mask-guided module are initialized by the MSRA normal distribution method. We fine-tune the model using SGD with $0.9$ momentum, $0.0001$ weight decay. The proposed PedHunter is trained on $16$ GTX 1080Ti GPUs with a mini-batch $2$ per GPU for CrowdHuman, Caltech-USA and SUR-PED, and the mini-batch size for Citypersons is $1$ per GPU. Multi-scale training and testing are not applied to ensure fair comparisons with previous methods. The specific settings of training process for different datasets are described in next section.

{\flushleft \textbf{Evaluation Metric.} }
The log-average miss rate over $9$ points ranging from $10^{-2}$ to $10^0$ FPPI (\ie, $\text{MR}^{-2}$) is used to evaluate the performance of the detectors (lower score indicates better performance). We report the detection performance for instances in full-body categories.

{\flushleft \textbf{Inference.} }
During inference, the mask-guided module is not used so that our PedHunter keeps the same computational cost as the baseline. We apply the Non-Maximum Suppression (NMS) with threshold $0.5$ to generate the top $100$ high confident detections per image as final results.

\begin{table*}[t]
\centering
\caption{Ablation experiments using $\text{MR}^{-2}$ (lower score indicates better performance). For CityPersons and Caltech datasets, we only report results on the Reasonable set.}
\setlength{\tabcolsep}{5.7pt}
\begin{tabular}{c|ccc|c|c|c|c|c}
\toprule[1.5pt]
{\multirow{2}{*}{Method}} &Strict &{\multirow{2}{*}{Mask-guided}} &Occlusion &{\multirow{2}{*}{Backbone}} &{\multirow{2}{*}{CityPersons}} &{\multirow{2}{*}{Caltech}} &{\multirow{2}{*}{CrowdHuman}} &{\multirow{2}{*}{SUR-PED}} \\
{} &Criterion &{} &-simulated &{} &{} &{} &{} &{} \\
\hline
Baseline &  & & &ResNet-50 &11.67 &3.26 &46.8 &57.9  \\
\hline
\multirow{3}{*}{PedHunter} &$\surd$ & & &ResNet-50 &10.59 &2.91 &45.2 &56.7 \\
&$\surd$ &$\surd$ & &ResNet-50 &9.62 &2.64 &43.6 &55.7 \\
&$\surd$ &$\surd$ &$\surd$ &ResNet-50 &8.32 &2.31 &39.5 &53.6 \\
\bottomrule[1.5pt]
\end{tabular}
\label{tab:ablation}
\end{table*}

\section{Datasets}
\label{sec:dataset}
In this section, we first introduce the proposed SUR-PED dataset, then present existing datasets including Caltech-USA, CityPersons and CrowdHuman datasets. 

\subsection{SUR-PED Dataset}
The proposed SUR-PED dataset is a benchmark dataset to better evaluate occluded pedestrian detectors in surveillance scenes. It contains $6,000$, $1,000$ and $3,000$ images for training, validation and testing subsets, respectively. The maximum resolution is $1920\times1080$. Images are crawled using some keywords of surveillance scenes from multiple image search engines including Google, Bing and Baidu. There are totally $162k$ human instances and an average of $16.2$ pedestrians per image with various kinds of occlusions. Each human instance is annotated manually with a head bounding-box, a human visible-region bounding-box and a human full-body bounding-box as in CrowdHuman. Different from the bounding boxes with fixed aspect ratio in CityPersons and Caltech, our annotation protocol is more flexible in real world scenarios where have various human poses. Figure \ref{fig:sur-ped} illustrates some examples from SUR-PED dataset, with the manner of annotations and the diversity of occlusion. In this paper, we report the detection performance for instances in pedestrian full-body categories. This dataset and a comprehensive evaluation tool will be released further. During training and inference on the proposed dataset, the input images are resized to have a short side of $800$ pixels as well as the long edges should be no more than $1333$ pixels. Due to the various aspect ratios of ground truths, the anchor aspect ratio setting adopts $0.5$, $1$ and $2$. For the first $13$ training epochs, the learning rate is set to $0.04$, and we decrease it by a factor of $10$ and $100$ for another $4$ and $3$ epochs, respectively.

\subsection{Caltech-USA Dataset}
The Caltech-USA dataset is one of the most popular and challenging datasets for pedestrian detection, which comes from approximately $10$ hours $30$Hz VGA video recorded by a car traversing the streets in the greater Los Angeles metropolitan area. The training and testing sets contain $42,782$ and $4,024$ frames, respectively. The commonly used $10\times$ training annotations \cite{DBLP:conf/cvpr/ZhangBOHS16} of Caltech-USA are refined automatically with only $16,376$ poor-quality instances in the training set. In our previous work, we re-annotate the dataset manually following the labeling rule and method in CrowdHuman as the extended version. We train the proposed method using $2\times$ scale of the image size with $2.44$ as the anchor aspect ratio. The initial learning rate is $0.04$ for the first $4$ epochs, and is reduced by $10$ and $100$ times for another $2$ and $1$ epochs.

\subsection{CityPersons Dataset}
The CityPersons dataset is recorded across $18$ different cities in Germany with $3$ different seasons and various weather conditions. The dataset includes $5,000$ images ($2,975$ for training, $500$ for validation, and $1,525$ for testing) with $\sim35,000$ manually annotated persons plus $\sim13,000$ ignored annotations. In our previous work, we additionally label the corresponding head bounding box for each annotated pedestrian instance as the extended version. The proposed PedHunter detector is trained on the training set and evaluated on the validation set. We enlarge input images by $1.3$ times and only use $2.44$ anchor aspect ratio for training. The initial learning rate is set to $0.02$ for the first $26$ epochs, and is decreased to $0.002$ and $0.0002$ for another $9$ and $5$ epochs.

\subsection{CrowdHuman Dataset} 
The CrowdHuman dataset is a benchmark dataset to evaluate pedestrian detectors in crowd scenarios. It is divided into training ($15,000$ images), validation ($4,370$ images) and testing ($5,000$ images) subsets. In particular, there are totally $470k$ human instances from the training and validation subsets, and $22.6$ persons per image with various kinds of occlusions. Each human instance is annotated with a head bounding-box, a human visible-region bounding-box and a human full-body bounding-box. Since the online evaluation server for the testing subset is not available until now, we train the proposed models on the CrowdHuman training subset and evaluate on the validation subset. During training, the input images are resized so that their short edges are at $800$ pixels while the long edges should be no more than $1333$ pixels at the same time. The anchor aspect ratios are set to $0.5$, $1$ and $2$ for CrowdHuman dataset. We train PedHunter with the initial learning rate $0.04$ for the first $16$ epochs, and decay it by $10$ and $100$ times for another $6$ and $3$ epochs.

\section{Experiments}
In this section, we first analyze the effectiveness of the proposed method, then evaluate the final model on common pedestrian detection benchmark datasets.

\subsection{Model Analyse}
To sufficiently verify the effectiveness of the proposed components, we construct ablation study on all four datasets including CrowdHuman, CityPersons, Caltech and SUR-PED. We first construct a baseline detector based on FPN \cite{DBLP:conf/cvpr/LinDGHHB17} with ResNet-50 \cite{DBLP:conf/cvpr/HeZRS16}. The performance of the baseline model is shown in Table \ref{tab:ablation}. For the CityPersons dataset, it obtains $11.67\%$ $\text{MR}^{-2}$ on the Reasonable set, outperforming the adapted FasterRCNN baseline in CityPersons by $1.13\%$. For the Caltech dataset, it achieves $3.26\%$ $\text{MR}^{-2}$ on the Reasonable set, which already surpasses all the state-of-the-art methods. For the CrowdHuman dataset, our implemented FPN baseline gets $46.8\%$ $\text{MR}^{-2}$ on the validation set, which is $3.62\%$ better than the reported baseline in \cite{DBLP:journals/corr/abs-1805-00123}. Thus, our baseline models are strong enough to verify the effectiveness of the proposed components. From the baseline model, we gradually apply the proposed strict classification criterion, mask-guided module and occlusion-simulated data augmentation method to verify their effectiveness. The parameter setting of all detectors in both training and testing is consistent for the fair comparison.

\subsubsection*{Strict Classification Criterion}
We first apply strict classification criterion onto baseline detectors to demonstrate its effectiveness. Comparing the detection results in Table \ref{tab:ablation}, we find that using the newly proposed strict classification criterion is able to reduce the $\text{MR}^{-2}$ by $1.08\%$ from $11.67\%$ to $10.59\%$ on the CityPersons dataset, by $0.35\%$ from $3.26\%$ to $2.91\%$ on the Caltech dataset, by $1.6\%$ from $46.8\%$ to $45.2\%$ on the CrowdHuman dataset, and by $1.2\%$ from $57.9\%$ to $56.7\%$ on the proposed dataset. We also provide visualization comparison of positive samples between whether applying the strict classification criterion or not in Figure \ref{fig:proposal}. It is obvious that the strict classification criterion effectively improves the quality of the positive samples in the R-CNN stage. Both performance improvements on four datasets and the visualization comparison demonstrate that strict classification criterion is effective for detecting the pedestrians in crowded scenes.

\subsubsection*{Mask-guided Module}
To validate the effectiveness of the mask-guided module, we add it after applying strict classification criterion. The ablation results are shown in Table \ref{tab:ablation}. Adding mask-guided module decreases the $\text{MR}^{-2}$ from $10.59\%$ to $9.62\%$ with $0.97\%$ improvement for the CityPersons dataset, from $2.91\%$ to $2.64\%$ with $0.27\%$ improvement for the Caltech dataset, from $45.2\%$ to $43.6\%$ with $1.6\%$ improvement for the CrowdHuman dataset, and from $56.7\%$ to $55.7\%$ with $1.0\%$ improvement for the proposed dataset. These improvements fully demonstrate the effectiveness of the proposed mask-guided module in occluded pedestrian detection. Meanwhile, we also present visualization comparison of feature maps between whether applying mask-guided module or not in Figure \ref{fig:feat-vis}. It is relatively obvious that mask-guided module contributes to more aggregate and strong response to pedestrian instances.

\subsubsection*{Occlusion-simulated Data Augmentation}
We propose the occlusion-simulated data augmentation to enrich the pattern and quantity of occlusion cases during training for stronger occlusion robustness. As shown in Table \ref{tab:ablation}, when we add this proposed augmentation strategy after the previous two improvements, the $\text{MR}^{-2}$ is further reduced by $1.3\%$ ($8.32\%$ \emph{vs.} $9.62\%$) on the CityPersons dataset, by $0.33\%$ ($2.31\%$ \emph{vs.} $2.64\%$) on the Caltech dataset, by $4.1\%$ ($39.5\%$\emph{vs.} $43.6\%$) on the CrowdHuman dataset, and by $2.1\%$ ($53.6\%$ \emph{vs.} $55.7\%$) on the proposed SUR-PED dataset. These comprehensive improvements indicate the effectiveness of the presented occlusion-simulated data augmentation for occluded pedestrian detection in a crowd.

\subsubsection*{Occlusion Subset Performance}
We also report the results on the heavy occlusion subset of CityPersons and Caltech-USA datasets in Table \ref{tab:ablation-occ}. After gradually applying the proposed three contributions, the MR$^{-2}$ performances are significantly improved by $6.21\%$ and $8.05\%$, respectively. Besides, our method shows a large margin over state-of-the-art methods, \ie, RepLoss and OR-CNN. These results on heavy occlusion subset demonstrate the effectiveness of the proposed PedHunter method.

\begin{table}[t]
\centering
\caption{MR$^{-2}$ performance on heavy occlusion subset.}
\footnotesize 
\setlength{\tabcolsep}{2.8pt}
\begin{tabular}{ccc|c|c}
\toprule[1.5pt]
Strict &{\multirow{2}{*}{Mask-guided}} &Occlusion  &{\multirow{2}{*}{CityPersons}} &{\multirow{2}{*}{Caltech-USA}} \\
Criterion &{} &-simulated &{} &{} \\
\hline
& & &49.74 &53.26  \\
$\surd$ & & &47.44 &51.10 \\
$\surd$ &$\surd$ & &46.33 &48.64 \\
$\surd$ &$\surd$ &$\surd$ &\textbf{43.53} &\textbf{45.31} \\
\hline
&RepLoss & &55.30 &63.36 \\
&OR-CNN & &51.30 &69.57 \\
\bottomrule[1.5pt]
\end{tabular}
\label{tab:ablation-occ}
\end{table}

\subsection{Benchmark Evaluation}

{\noindent \textbf{CityPersons.} }
We compare PedHunter with TLL (MRF) \cite{Song_2018_ECCV}, Adapted FasterRCNN \cite{DBLP:conf/cvpr/ZhangBS17}, ALFNet \cite{Liu_2018_ECCV}, Repulsion Loss \cite{DBLP:conf/cvpr/WangXJSSS18}, PODE+RPN \cite{DBLP:conf/eccv/ZhouY18} and OR-CNN \cite{DBLP:conf/eccv/ZhangWBLL18} on the CityPersons validation subset in Table \ref{tab:cityperson-val}. Similar with previous works, we evaluate the final model on the Reasonable subset of the CityPersons dataset. The proposed PedHunter method surpasses all published methods and reduces the $\text{MR}^{-2}$ score of state-of-the-art results from $11.0\%$ to $8.32\%$ with $2.68\%$ improvement compared with the second best method \cite{DBLP:conf/eccv/ZhangWBLL18}, demonstrating the superiority of the proposed method in pedestrian detection.

\begin{table}[h]
\centering
\caption{$\text{MR}^{-2}$ performance on the CityPersons validation set. Scale indicates the enlarge number of original images.}
\small
\setlength{\tabcolsep}{7.2pt}
\begin{tabular}{c|c|c|c}
\toprule[1.5pt]
Method &Backbone &Scale &{\em Reasonable} \\
\hline
TLL (MRF) & ResNet-50 &- &14.40 \\
Adapted FasterRCNN & VGG-16 & $\times1.3$ & 12.97 \\
ALFNet & VGG-16 &$\times$1 &12.00 \\
Repulsion Loss & ResNet-50 &$\times$1.3 &11.60 \\
PODE+RPN & VGG-16 &- &11.24 \\
OR-CNN & VGG-16 &$\times$1.3 &11.00 \\
\hline
PedHunter & ResNet-50 &\textbf{$\times$1.3} &\textbf{8.32} \\
\bottomrule[1.5pt]
\end{tabular}
\label{tab:cityperson-val}
\end{table}

\begin{figure}[t]
\centering
\includegraphics[width=0.95\linewidth]{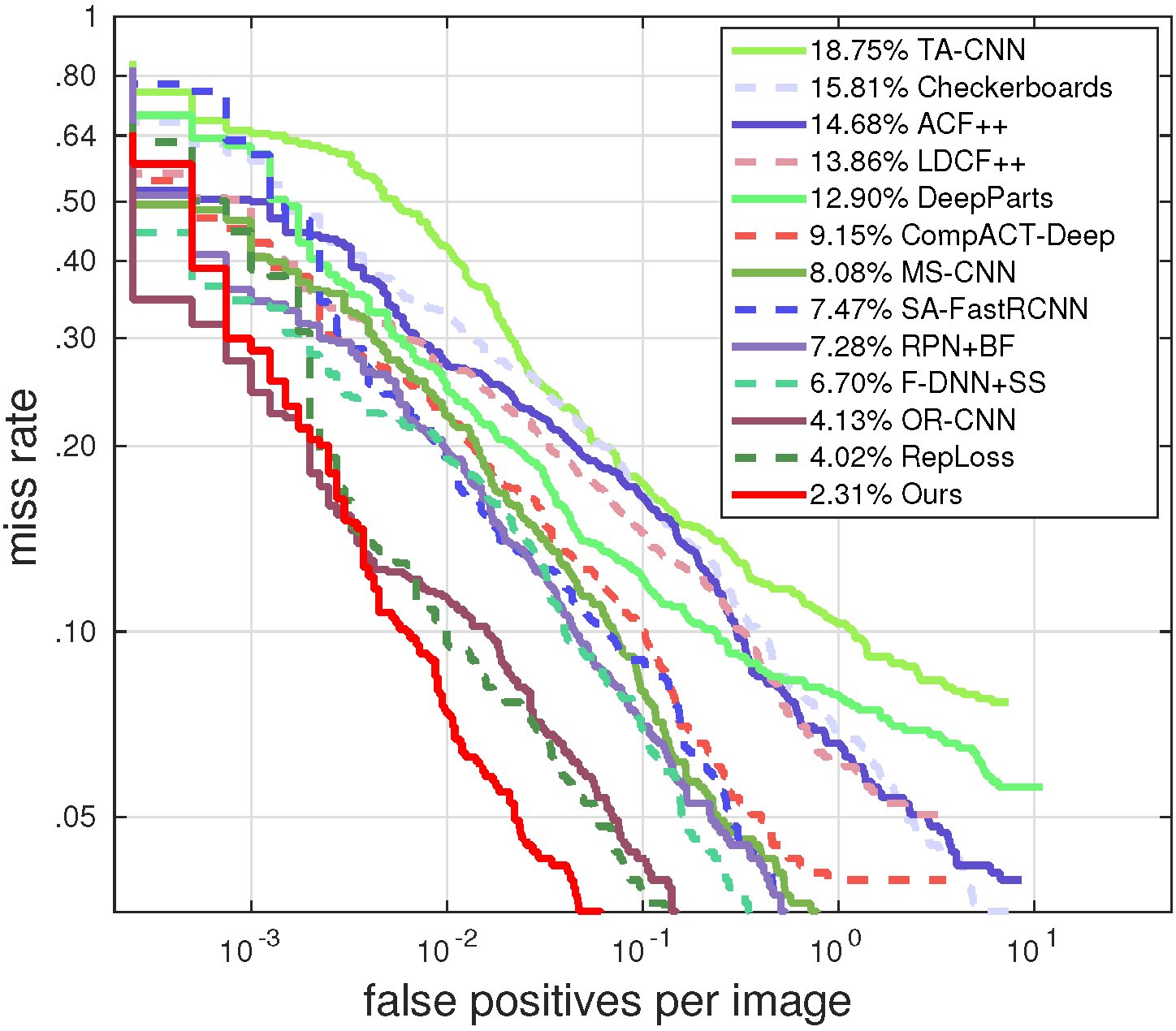}
\caption{${\text{MR}}^{-2}$ scores of different state-of-the-art methods on the Caltech-USA dataset.}
\label{fig:caltech}
\end{figure}

{\noindent \textbf{Caltech-USA.} }
\label{caltech-res}
Figure \ref{fig:caltech} shows the comparison of the PedHunter method with other state-of-the-art methods on the Caltech-USA testing set. All the reported results are evaluated on the widely-used Reasonable subset, which only contains the pedestrians with at least $50$ pixels tall and occlusion ratio less than $35\%$. The proposed method outperforms all other state-of-the-art methods by producing $2.31\%$ $\text{MR}^{-2}$.

\section{Conclusion}

This paper presents an efficient method to improve the occluded pedestrian detection accuracy in crowded scenes. Specifically, a mask-guided module is designed to enhance the representation and discrimination of features. Meanwhile, a strict classification criterion is introduced to eliminate common false positives in crowded scenes. Moreover, an occlusion-simulated data augmentation method is proposed to improve the robustness of network against occlusion. Besides, we collect a new occluded pedestrian detection benchmark dataset in surveillance scenes. Consequently, we achieve state-of-the-art performances on common pedestrian detection datasets. The proposed dataset, source codes and trained models will be public to facilitate further studies of occluded pedestrian detection.

\clearpage

\bibliographystyle{aaai}
\bibliography{reference}
\end{document}